\pdfoutput=1
\documentclass[11pt]{article}

\usepackage{EMNLP2023}

\usepackage{times}
\usepackage{latexsym}

\usepackage[T1]{fontenc}
\usepackage[utf8]{inputenc}

\usepackage{microtype}

\usepackage{inconsolata}
\usepackage{graphicx}
\usepackage{booktabs}
\usepackage{paralist}

\usepackage{stfloats}  % Load the stfloats package

\title{On the effect of curriculum learning with developmental data\\ for grammar acquisition}

\author{%
  Mattia Opper\textsuperscript{\,a}
  \and J. Morrison\textsuperscript{\,a}
  \and N. Siddharth\textsuperscript{\,a,b} \\[0.5ex]
  \textsuperscript{a} University of Edinburgh;\;
  \textsuperscript{b} The Alan Turing Institute \\[0.5ex]
  \texttt{\{m.opper,j.morrison,n.siddharth\}@ed.ac.uk}
}

\begin{document}
\maketitle
\begin{abstract}
This work explores the degree to which grammar acquisition is driven by language `simplicity' and the source modality (speech vs. text) of data.
Using BabyBERTa \cite{babyberta} as a probe, we find that grammar acquisition is largely driven by exposure to speech data, and in particular through exposure to two of the BabyLM \cite{BabyLM} training corpora: AO-Childes and Open Subtitles.
We arrive at this finding by examining various ways of presenting input data to our model.
First, we assess the impact of various sequence-level complexity based curricula.
We then examine the impact of learning over `blocks'---covering spans of text that are balanced for the number of tokens in each of the source corpora (rather than number of lines).
Finally, we explore curricula that vary the degree to which the model is exposed to different corpora.
In all cases, we find that over-exposure to AO-Childes and Open Subtitles significantly drives performance.
We verify these findings through a comparable control dataset in which exposure to these corpora, and speech more generally, is limited by design.
Our findings indicate that it is not the proportion of tokens occupied by high-utility data that aids acquisition, but rather the proportion of training steps assigned to such data.
We hope this encourages future research into the use of more developmentally plausible linguistic data (which tends to be more scarce) to augment general purpose pre-training regimes.
\end{abstract}

\section{Introduction}
% points to hit in the intro: developmentally plausible pre-training, acquisition of grammatical knowledge, resource requirements
% not sure how much it is necessary to talk about the BabyLM training data and the background behind that

% the problem
Pre-training modern LLMs has become an increasingly resource intensive process, often requiring hundreds of GPU hours, and enough electricity to power a small village. These requirements have led to model creation increasingly becoming restricted to the few actors that are able to muster the resources necessary, excluding many from being able to participate in researching the field.

% prior contributions towards the solution
On the other hand, recent work  \cite{babyberta, PlantTrees} has shown that Transformer LLMs can acquire knowledge of grammar and syntax with less data scale than was previously thought necessary, provided that they are exposed to simpler forms of language. These findings provide a hope that research on pre-training can once again become accessible to the community as a whole.

% even if you don't need loads of data you still get better performance with greater model complexity and more steps

However, even if scale may not be such a strict requirement for the acquisition of linguistic knowledge, there are two tendencies exhibited by transformer models that may still be barriers to accessibility. Firstly, simply increasing the number of training steps generally yields better results. In fact, recent work by \citet{MurtyGrokking} has shown that continuing training long past \textit{train loss saturation} can lead to acquisition of a bias towards tree-likeness. While a fascinating finding in its own right (as hierarchical structure is considered a central feature of natural language) many groups simply won't have the GPU hours necessary to reach this point, so resources may remain a barrier. Secondly, it is often the case that simply increasing the complexity of a model can be beneficial (e.g. greater depth can aid syntactic generalisation \cite{PlantTrees}), but increasing complexity also increases cost.

This work investigates whether we can use the starting small approach to curriculum learning \cite{Elman1993LearningAD} combined with a small scale developmentally plausible pre-training set to aid model grammar acquisition without necessitating an increased budget of training steps. Our findings are mixed. We were unable to significantly outperform a random sampling baseline over all the pre-training corpora contained in the strict-small track. However, we are able to attribute this to the prevalence of high-utility simple speech data. We demonstrate through the use of a control corpus that in a setting where this high-utility data is more scarce, the benefits of developmentally ordered learning start to show themselves.
% This work investigates which techniques can be applied to further reduce the cost of pre-training. Our aim was to investigate whether the variance in complexity between the BabyLM \cite{BabyLM} pre-training corpora can be utilised to form a structured training regimen. Our findings are mixed. We were unable to outperform a random sampling baseline over all the pre-training corpora, but are able to attribute this to the prevalence of high-utility simple speech data. We demonstrate that in a setting where this data is more scarce, a complexity based curriculum can prove beneficial for grammar acquisition.

\section{Related Work}
% just gather up the papers first
\citet{Elman1993LearningAD}'s seminal early work presented the idea of starting small, whereby a model is first exposed to simpler data before moving on to more complex types of input. The idea is that complex data might get the model to learn `false friend' heuristics that are actually harmful in the long run, but simple data might get it to learn in a way that generalises well. However, this hypothesis is not without controversy. \citet{Rohde1999LanguageAI} found that networks trained on complex sentences from the start performed better than those trained on simpler sentences initially, contradicting the starting-small hypothesis. They argue that previous studies supporting the starting small hypothesis may have terminated the training of complex networks too early. \citet{BengioC} train a language model using a curriculum learning strategy where only spans of text containing the first 5k most frequent words are included, then expanding to the first 10k and so on. They find that while a random sampling baseline initially achieves a superior loss, with sufficient updates the curriculum strategy comes to a better minimum and converges more stably.

These approaches have in common that they gradually reveal more and more of the dataset. An alternative approach is a single-phase curriculum where the input data is sorted by some criterion and then presented to the model in a fixed ordering. The model goes through the curriculum once, and does not revisit simpler data once it transitions to more complex data. The success of the single phase approach depends heavily on how complexity is defined, and has shown dubious results when applied to NLP \cite{NoLingCurriculum, surkov-etal-2022-data}. Even under a developmentally plausible setting, the efficacy of the single phase approach has been shown to be mixed \cite{babyberta}.

\section{BabyBERTa}
\subsection{Model and Training Details}
% introduce babyberta need to add a discussion regarding unmasking also
The baseline model architecture we use in this work is an adaptation of BabyBERTa \cite{babyberta}. BabyBERTa is a variant of RoBERTa \cite{RoBERTa}, with a few key differences:

\begin{compactdesc}
    \item[No Unmasking:] RoBERTa had used unmasking to minimise the disparity between pre-training and fine-tuning (where no mask tokens are used). Instead, BabyBERTa prioritises the finding that removing unmasking substantially improves model grammar acquisition.
    \item[No length truncation:] Sequences which exceed the max length set in BabyBERTa are excluded instead of truncated. This ensures the model is only provided with whole utterances that correspond to a coherent linguistic unit.
    \item[Smaller Size:] BabyBERTa is both shallower (fewer layers) and narrower (lower hidden and feed-forward size) than the original RoBERTa.
    \item[Training Data and Vocab Size:] BabyBERTa is pre-trained on child directed speech and uses a substantially smaller vocabulary size in order to mimic that of a 6-year-old (theorised to be roughly 6k words).
\end{compactdesc}

\noindent%
We adopt this architecture for use in our paper with some alterations:

\begin{compactdesc}
    \item[Increased Vocabulary:] The BabyLM training corpora consist of more diverse data than AO-Childes, and encompass a wider range of developmental complexity. Consequently, a greater vocabulary size may be beneficial. We performed a grid search over vocabulary sizes 10k, 20k, 30k, 40k and 50k and found 30k to be optimal.
    \item[Increased Width:] We double the hidden size and feed-forward network dimension of the original BabyBERTa from 256 to 512 and 1024 to 2048 respectively. These changes yielded slight improvements in BLiMP performance, but without them the model performed substantially worse on NLI tasks than the RoBERTa baseline provided for the challenge. However, increased width yields only minimal improvements in terms of grammar acquisition. We tested increasing the depth of the model (more layers), but found this yielded no improvements within the pre-training step budget we had available, neither did increasing the number of attention heads.
\end{compactdesc}

Our remaining model parameters are the defaults for RoBERTa from the transformer's library \cite{transformers}. We use relative key query positional embeddings and set our max sequence length during training to 128 for efficiency reasons, and follow the no-truncation strategy. We set the learning rate to 1e-5 and the max number of steps to 120k using batch size 128. Unless stated otherwise, all our experiments utilise these same hyper-parameters. We utilise dynamic masking as with the original RoBERTa, and no unmasking following BabyBERTa in all cases without exception. While the latter choice may impact downstream performance in the fine-tuning tasks, the focus of this paper is largely on grammar acquisition as measured by the zero-shot evaluation suite and here removing unmasking proved beneficial.

\section{Sequence Complexity Curricula}
Our first point of investigation was to examine whether we could use sequence complexity based curricula to improve grammar acquisition. In the original BabyBERTa paper, the authors found that training on AO-Childes in its original ordering (which corresponds to age ordering, hence AO) led to better grammar acquisition than the reverse, but failed to outperform a random sampling baseline. They attribute this failure to a lack of vocabulary diversity in each batch when using age ordering. By contrast, the BabyLM pre-training corpora exhibit varying complexities (AO-Childes or Open Subtitles are on average much simpler than Wikipedia, see Figure~\ref{fig: corpheatmap}), as well as variance in complexity within the corpora. Consequently, we hypothesised that we may be able to scaffold learning by presenting sequences to the model in order of complexity, while mitigating the potential issue of vocabulary and domain diversity by drawing these sequences from across all the source corpora.

% % talk about the metrics
% What is it that sets the speech data apart from text? To analyse this, we computed various sequence complexity metrics for each of the different corpora in order to identify any trends (shown in Figure Add Fig).  The speech corpora in general have a far lower sequence length, token to type ratio, word length, bigram diversity and entropy, than their textual counterparts. Effectively meaning that they generally exhibit simpler and highly regular language. It is worth noting that this is influenced by the fact that the text corpora also represent more complex units like articles, which will skew some metrics, even if the sentence wise content of a given text corpus may be simple. \

% % why might we want a more complex curriculum
% If the cause for speech data being useful for learning grammar is its simplicity, it is worth investigating whether we can somehow capture this notion and create a more structured curriculum this way. The speech corpora are not uniform in their complexity, and neither are the sequences contained within each individually. Simply training on speech first or just the concatenation of all corpora means that there is no control over when the model encounters complex vs. simple sequences. Instead, it relies on the balance of the dataset being such that simpler sequences are encountered more frequently. This raises the question: what happens if we try to control training so that simpler sequences are deliberately introduced first. Does this establish a solid learning foundation for our model?

\subsection{Curriculum Types}
We tested three kinds of curricula using different measures for complexity. As we were submitting to the strict small track, we only used sequence complexity metrics that could be easily inferred from the raw data.
We call lines of the corpora `sequences' for lack of a better term. Each corresponds to a linguistically coherent unit, but they can vary from short transcribed utterances to full articles. It is likely that better curricula can be created by using more complex and linguistically motivated metrics, but without the use of external resources this is difficult to achieve. The three types we tried are:

\begin{compactdesc}
    \item[Entropy:] Entropy favours highly likely sequences, but penalises based on length. This should order data such that the most likely shortest sequences appear first, allowing the model to learn simple local dependencies before moving to more complex data.
    \item[Unigram Probability:] Orders sequences by the average unigram probability of their tokens. This is similar to entropy, except without penalising length directly. The idea here is that the model can learn good representations for highly likely tokens first and use that to inform its decision around more complicated/rarer tokens later down the line. The approach is similar to that of \citet{BengioC}.
    \item[Block:] Introduced by \citet{BlockCurriculum} in the block curriculum, block size is increased during the course of training. This allows the model to first learn to optimise local dependencies before moving to longer range ones. The block curriculum differs from the other two in that each stage of learning does not present a subset of sequences, but rather is over the entirety of data in all the corpora, with each stage providing a greater context window for the model to consider. Secondly, by utilising blocks, each input consists of a span of tokens rather than a linguistically coherent unit like a transcribed utterance or article, and can include segments that represent partial units both at the start or end of a block. This means that the model must learn to identify the boundaries between coherent units during training, which may be a burden.
\end{compactdesc}

\subsection{Creation}
% entropy and unigram curriculum
We first tokenised all sequences using the model's tokeniser, then calculated probabilities for each token using MLE, and scored each sequence, and subsequently re-ranked the data. The re-ranked sequence were then divided into different stages, by chunking according to rank. We used 4 stages for all curricula, with each stage containing a roughly equal number of sequences. Increasing this number did not yield significant improvements.

% block curriculum
In the original block curriculum \citet{BlockCurriculum} use block sizes 64, 128, 256 and 512, with the maximum batch size that could fit on their GPU at each step. We adopt this approach, but following initial findings that significantly smaller block sizes proved more beneficial than larger ones (potentially as a result of us limiting the max number of steps to 120k to enforce consistency across experiments), we instead switched to block sizes 16, 32, 64, 128.

% Using concatenation
In some preliminary training runs, we tested both the single phase and starting small approaches to curriculum learning. The single phase approach proved significantly inferior and exhibited a tendency towards catastrophic forgetting. Instead, we used the following strategy: Each stage introduces new data for training, and the model is trained on the data in the current stage concatenated with that of all stages seen prior. This approach worked best for us. Each stage was trained on for 30k steps, totalling a combined 120k. As a baseline, we trained using random sampling over the whole data, also for 120k steps.

\subsection{Summary}
\begin{figure}[t]
\centering
\includegraphics[width=\columnwidth]{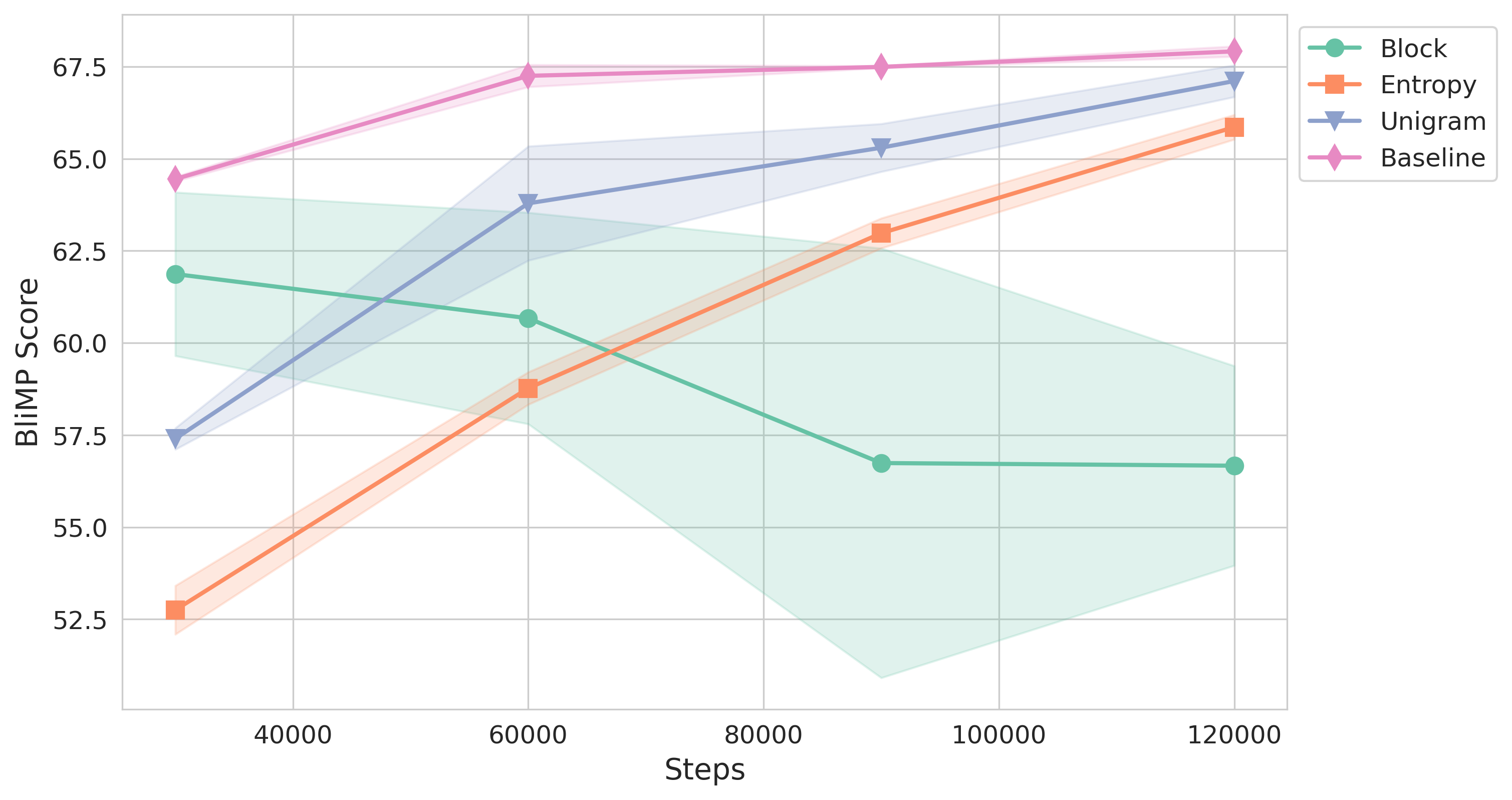}
\caption{Zero-shot performance for curricula vs. random-sampling baseline with training (over 3 seeds).}
\label{fig: cscore}
\end{figure}
%J: Figure 7 shows curriculum better in everything except Npi Licensing, but Figure 1 Blimp score worse, so blimp not good indicator of relative performance in these Fig 7 tasks?
%M: These are different curricula under different settings. Should probably be explained in the captions if that’s confusing.
%J: In one case we use 'baseline', in the other 'random', so it's a bit confusing to me because we previously described 'baseline' as being random.

Figure~\ref{fig: cscore} shows results. None of the curricula were able to outperform a baseline measure of simply sampling random sequences from the concatenation of all the datasets. Though the sequence complexity based curricula showed improvement throughout training, the block curriculum got worse with each stage. This raised two follow-up questions for us. First, what causes the random sampling baseline to do so well? Second, is using blocks as inputs rather than sequences causing the block curriculum to fail, or some other factor \footnote{The large variance exhibited by the block curriculum suggests significantly more steps would be needed to perform well.}?

\section{Investigating Random Sampling}
\begin{figure}[t]
\centering
\includegraphics[width=\columnwidth]{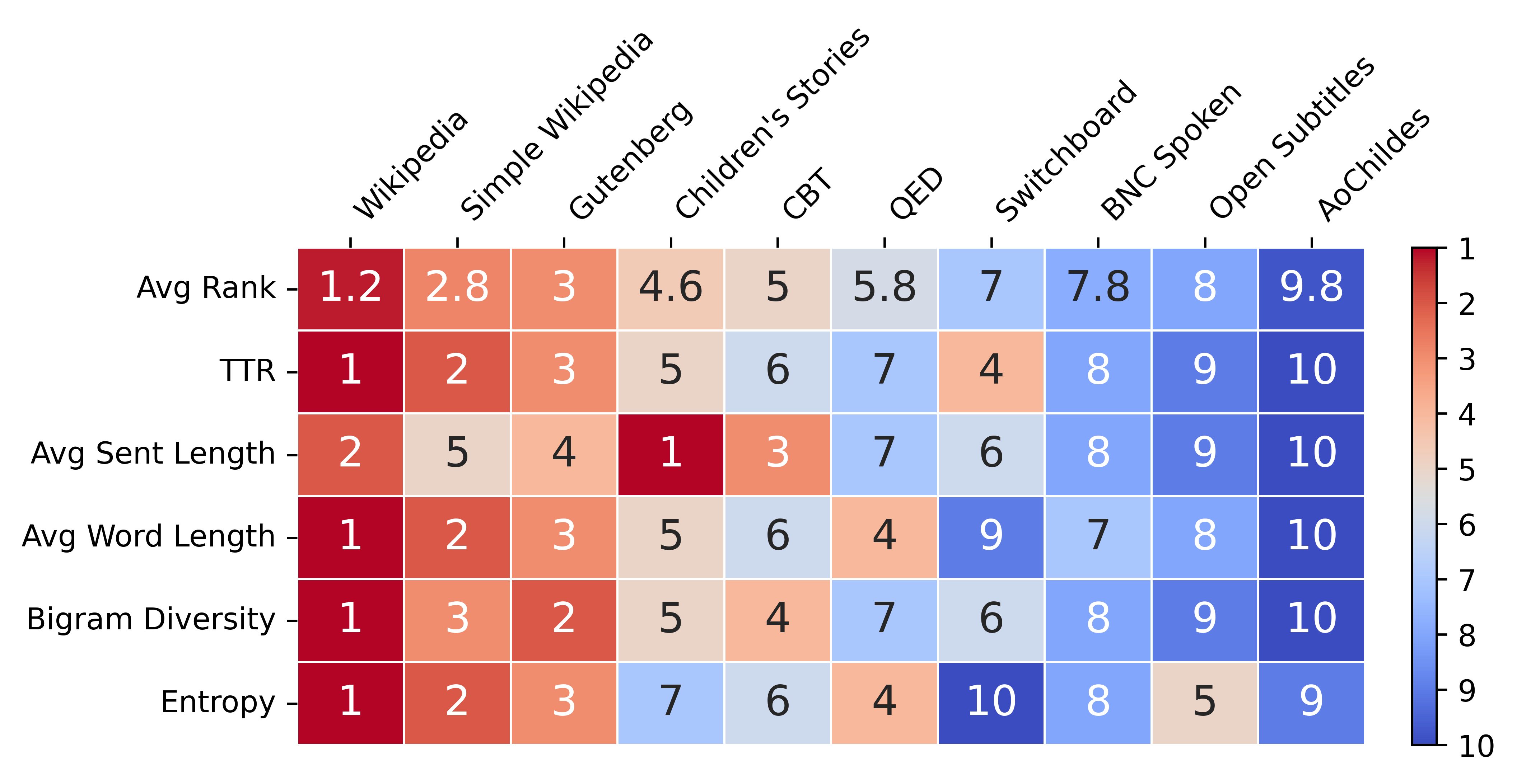}
\caption{Heatmap ranking of the BabyLM Strict Small training corpora according to complexity measures.}
\label{fig: corpheatmap}
\end{figure}

\begin{figure}[t]
\centering
\includegraphics[width=\columnwidth]{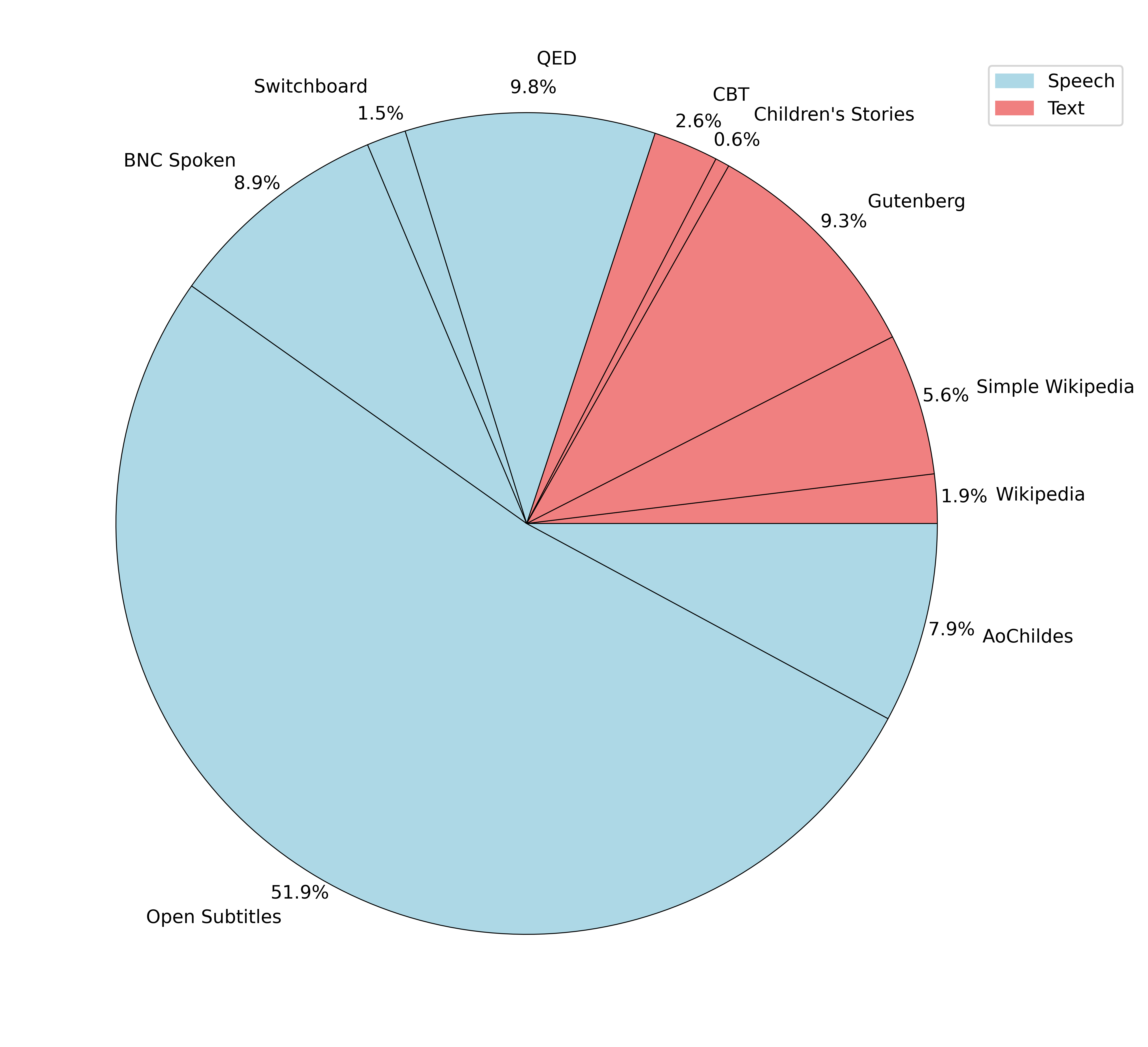}
\caption{Distribution of line counts across the ten language corpora, with each line treated as a unique sequence. The percentages represent the proportion of total lines that each individual corpus contributes to the overall dataset.}
\label{fig: seqprop}
\end{figure}

Why might random sampling be successful? Let us begin by examining how we present our data. In terms of number of tokens, the BabyLM pre-training corpora are roughly equally divided between the source modalities: text and (transcribed) speech. Though there is a slight weighting in favour of speech, which comprises 56\% of total tokens. Now let us contrast this with the relative complexity of each corpus (see Figure~\ref{fig: corpheatmap}). We can see that the speech corpora on average, across all metrics, contain far simpler language than the text corpora. Secondly, as we were submitting to the strict small track we do not perform any augmentation on the data, including sentence tokenisation. This means that the random sampling baseline takes as input lines from each corpus. If we examine the distribution of number of lines between corpora, we find a very different division compared with the number of tokens. Figure~\ref{fig: seqprop} shows the breakdown. Looking at the number of lines, the balance between transcribed speech and text data becomes highly unequal, with transcribed speech now comprising a total of 80\% of all examples. Secondly, the two corpora which contain on average the simplest language (AO-Childes and Open-Subtitles) represent 59.8\% of all lines, and these may be responsible for driving the majority of grammar acquisition. If this is the case, then it may explain the performance of the random sampling baseline, as it is more likely to see sequences from these two corpora than any others, while still being provided a degree of diverse examples in each batch. By contrast, when the input is treated as blocks rather than lines, the balance between speech and text inputs corresponds to the proportion of number of tokens. Alternatively, it may simply be that training on blocks requires more steps so that the model can identify linguistically coherent units.

To test this hypothesis, we train on both models, taking either blocks or lines from the corpora (henceforth referred to as sequences) as input. We train for an equal number of steps (120k). We report results for block size 32, as when trained for the full number of steps, this worked best out of all the variations tested in the block curriculum.

\subsection{Summary}
\begin{figure}[]
\centering
\includegraphics[width=\columnwidth]{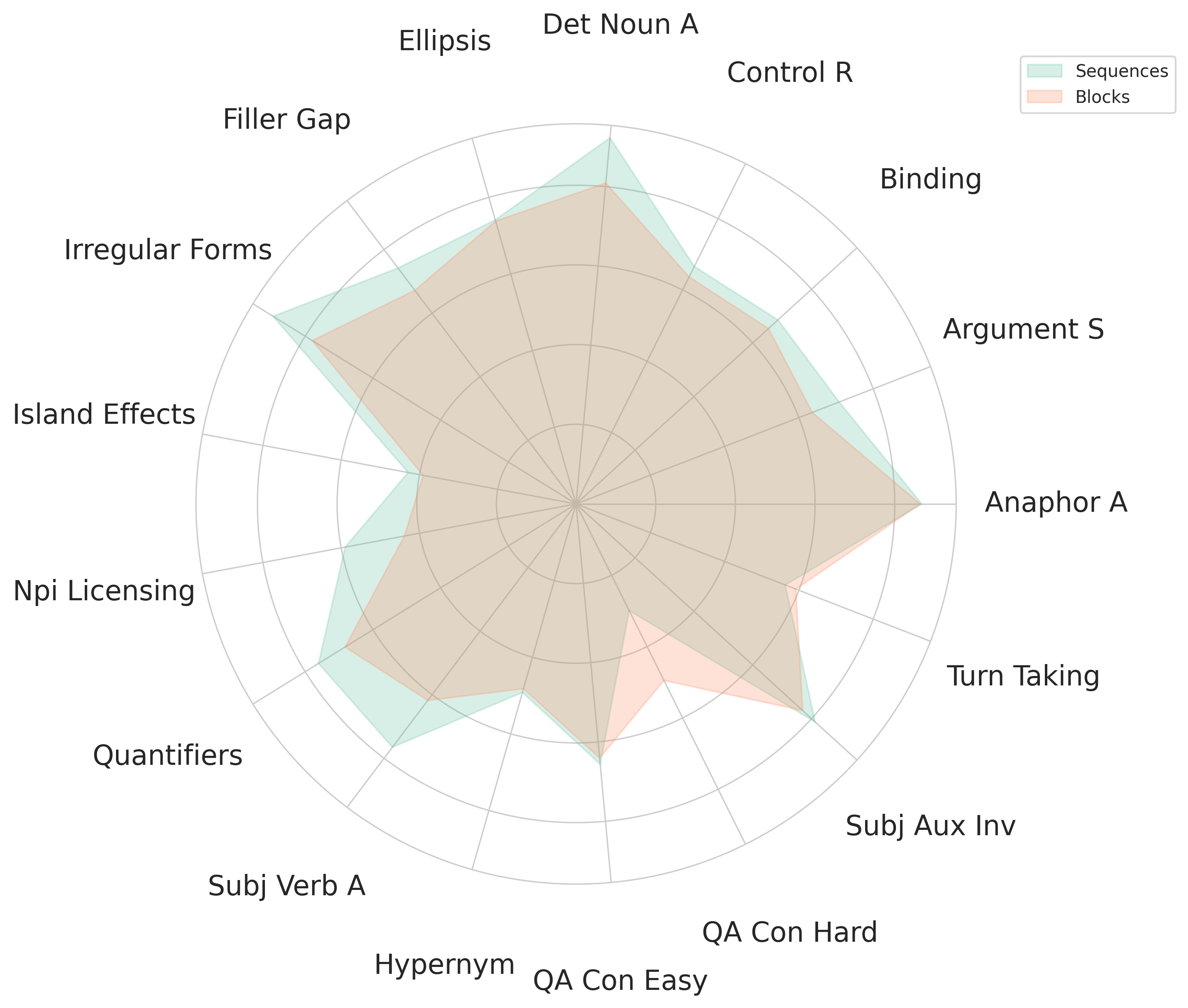}
\caption{By-task breakdown of zero-shot performance when input data is either a linguistically coherent sequence or a block. Results averaged over 3 seeds.}
\label{fig: seqvblock}
\end{figure}

Even when trained for a greater number of steps we find that sequences as input still quite substantially outperform blocks. Results are shown in Figure~\ref{fig: seqvblock} and Table~\ref{tab:seqvblock}. The only exception is on the held out tasks, however, this is due to the block variant of the model essentially having random accuracy on the QA congruence tasks (close to 50\%) while the sequences variants appear to have learned to solve the easy tasks, but fail at the hard ones (see Table~\ref{tab:seqvblocfull} for full results by for each task).

We can conclude from this that providing linguistically coherent units as input is beneficial for overall efficient grammar acquisition, despite the fact that the model is disproportionately being exposed to speech data, and therefore only a subset of the overall tokens throughout pre-training. However, we still need to disentangle whether it is speech that is driving this effect or the fact that the model is being presented linguistically coherent units.

\begin{table}[t]
\centering
%\resizebox{\columnwidth}{!}{%
\caption{By-task breakdown of zero-shot performance between models utilising random sampling strategies where inputs are either linguistically coherent sequences or blocks. Results averaged over 3 seeds.}
\label{tab:seqvblock}
\begin{tabular}{lll}
Tasks & Blocks & Sequences\\
\midrule
Original & 65.98 $\pm$ 1.02 & \textbf{73.11 $\pm$ 0.89} \\
Held Out  & \textbf{59.59 $\pm$ 0.6 } & 56.45 $\pm$ 0.88 \\
Overall & 64.1 $\pm$ 0.2 & \textbf{68.21 $\pm$ 0.23} \\
\end{tabular}%
%}
\end{table}

\begin{figure}[t]
\centering
\includegraphics[width=\columnwidth]{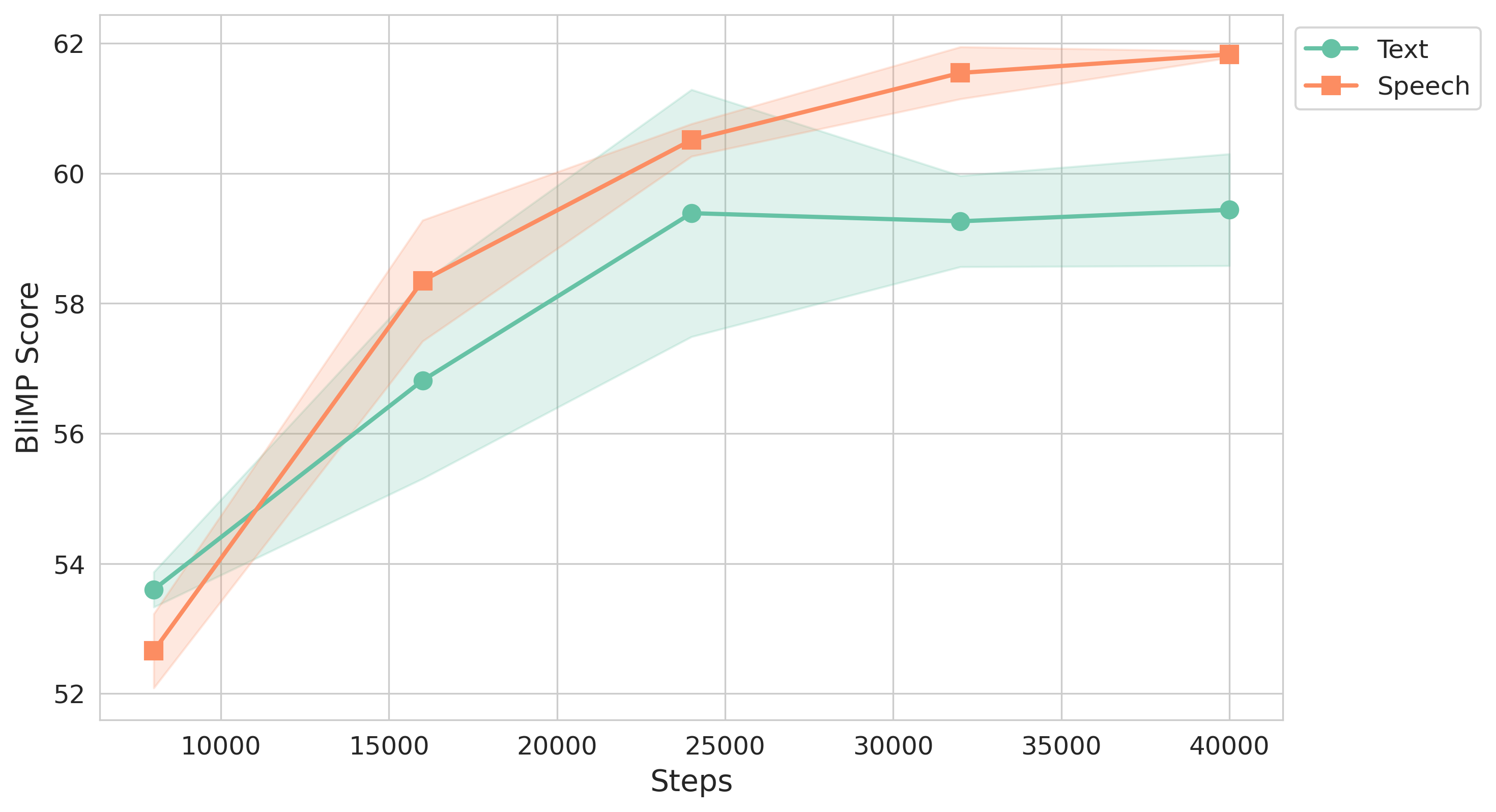}
\caption{Zero-shot performance by step when the model is trained on either the transcribed speech or text portions of the pre-training corpora (over 3 seeds).}
\label{fig: spvtxt}
\end{figure}

\section{Speech versus Text}
% introduce prior work on the topic
\subsection{Efficient Acquisition by Modality}
Prior work examining the impact of pre-training on AO-Childes \cite{babyberta, PlantTrees} has shown that utilising this simpler form of language enables more efficient acquisition of grammatical knowledge and encourages a bias towards hierarchical generalisation in transformer language models. So in our case, over exposing the model to simpler data such as speech may be driving performance. To test this, we perform two ablations. First, we compare the impact of training on only one source modality (for a reduced number of steps) to assess whether text-only or speech-only provides a better starting point for acquisition. This actually ought to favour the textual data in some respects because it contains longer sequences and therefore should provide more signal per step, as each input will contain more masks and contexts while still representing a linguistically coherent unit. Figure~\ref{fig: spvtxt} shows results for the first ablation comparing the two modalities when trained for 40k steps each. Training on transcribed speech consistently outperforms training on text alone, and leads to more stable improvements than just text, indicating that speech is a better starting point.

\subsection{Speech Data as a Foundation}
As a second follow-up investigation, we once again trained on two different settings. In the first we train on speech first and then the concatenation of text and speech for 60k steps respectively. This is to check whether we can build a foundation from speech data alone, and then transition to including both modalities. However, here text data only occupies 10\% of the overall proportion of inputs, and is only observed in the later stages of training. As a control, we also try the inverse, starting with text first and then transitioning to the concatenation of all the corpora. This means that the text data now provides 60\% of all the total inputs and speech is only introduced to the model later in training, no longer acting as a foundation. Results are in Table~\ref{tab:scaffcomp}. Weighting towards speech beats the text-first control in the original BLiMP tasks, and reaches similar performance to random sampling given their standard deviations overlap. On the held out tasks, the highly variable text-first results are sometimes competitive.

\label{sec: speechscaf}

% \begin{figure}[t]
% \centering
% \includegraphics[width=\columnwidth]{images/speechscaf.png}
% \caption{Your caption goes here}
% \label{fig: speechscaf}
% \end{figure}

\begin{table}[]
\centering
\caption{Comparison of Ordering Effects Given Source Modality. Results averaged across 3 seeds.}
\label{tab:scaffcomp}
\resizebox{\columnwidth}{!}{%
\begin{tabular}{llll}
Training Data & Original & Held Out & Overall\\
\midrule
Speech $\rightarrow$ Speech+Text & \textbf{72.99 $\pm$ 0.53} & \textbf{56.26 $\pm$ 1.3} & \textbf{67.74 $\pm$ 0.77} \\
Text $\rightarrow$ Speech+Text & 71.69 $\pm$ 0.6 & \textbf{53.77 $\pm$ 2.78} & 66.42 $\pm$ 1.2 \\
\midrule
Speech+Text & \textbf{73.11 $\pm$ 0.89} & \textbf{56.45 $\pm$ 0.88} & \textbf{68.21 $\pm$ 0.52} \\
\end{tabular}%
}
\end{table}

\subsection{Summary}
We find that transcribed speech leads to improved BLiMP performance and lower variance compared with text only data. Based on this finding, we investigated whether we could design a simple two stage curriculum where we first train the model on speech only and then transfer to the full dataset. Under this setting, performance is roughly equal to random sampling, and shows some very slight improvements compared to the text-first reverse curriculum. This is despite withholding $\approx$ 50\% of the total tokens (contained in the text portion of the data) until the latter half of training.

\section{Corpora Complexity Curricula}

% \begin{figure}[t]
% \centering
% \includegraphics[width=\columnwidth]{images/concatcurriculum.png}
% \caption{Your caption goes here}
% \label{fig: concat_c}
% \end{figure}

\begin{figure}[t]
\centering
\includegraphics[width=\columnwidth]{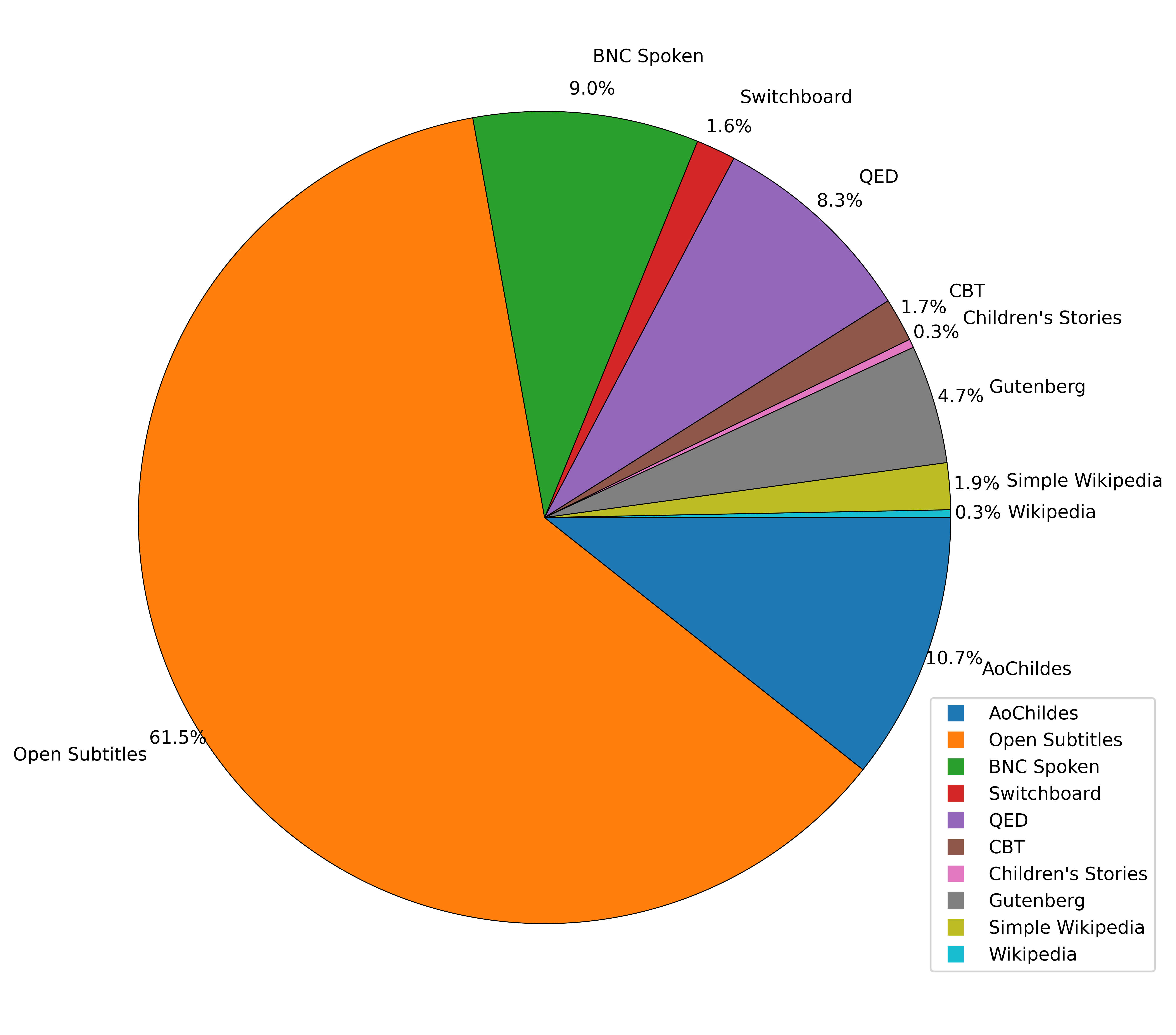}
\caption{Proportion of total inputs comprised by each of the corpora using the corpus complexity curriculum.}
%\caption{Proportion each corpus' data sequence is viewed when using the corpus complexity curriculum. The data is weighted by both the size of the corpus (number of lines in the range of 0 to 128 words) and the number of times it is used in different stages of the curriculum.}
%\caption{Percentage of times input data from each corpus is observed by the model using the corpus complexity curriculum.}
\label{fig: concat_c}
\end{figure}

Having found that speech data can provide a better foundation than text, and that over exposure may be behind the random sampling baselines performance, we conduct a follow-up investigation. How much exposure to more complex data is necessary in order to achieve grammar acquisition? To probe this question, we use the same strategy for our curriculum by training on a stage and the concatenation of all previous stages. This time we define our ordering using the average rank across our various corpus complexity measures as shown in Figure~\ref{fig: corpheatmap}. So our ordering starts with AO-Childes and ends with Wikipedia. The curriculum is simply the corpus complexity ordering, with two caveats. We treat BNC spoken and switchboard as one corpus, as switchboard is too small to warrant a new stage. We also do the same for CBT and children's stories, as they are very similar in terms of complexity. Using this form of curriculum further increases the model's exposure to simple data, with AO-Childes and Open Subtitles now representing 72.2\% of all total training examples, compared with 59.8\% before, and Wikipedia representing only 0.3\% (see Figure~\ref{fig: concat_c}). We again implement the reverse curriculum as a control measure, starting with Wikipedia and finishing with AO-Childes, and compare results to the random sampling baseline (see Table~\ref{tab: corpcomplex}). The simple to complex curriculum yields marginally better results overall compared to the random sampling baseline, and the gap with the reverse curriculum is wider here than for the previous speech versus text curricula.

\begin{table}[!b]
\centering
\caption{Comparison of performance by corpora complexity ordering. Results averaged across 3 seeds.}
\label{tab: corpcomplex}
\resizebox{\columnwidth}{!}{%
\begin{tabular}{llll}
Training Data & Original & Held Out & Overall\\
\midrule
Simple $\rightarrow$ Complex & \textbf{74.14 $\pm$ 0.39} & \textbf{55.9 $\pm$ 0.74} & \textbf{68.77 $\pm$ 0.04} \\
Complex $\rightarrow$ Simple & 71.89 $\pm$ 0.9 & \textbf{54.29 $\pm$ 2.74} & 66.72 $\pm$ 0.42\\
\midrule
Random Sampling & \textbf{73.11 $\pm$ 0.89} & \textbf{56.45 $\pm$ 0.88} & \textbf{68.21 $\pm$ 0.52} \\
\end{tabular}%
}
\end{table}

However, the marginality of the increase compared to the random sampling baseline makes it difficult to make any strong claims regarding the effect of ordering. We suspect this was because the BabyLM training data is already favourable for grammar acquisition and weighted towards speech, and expect we would observe greater benefits over random sampling in a setting where the data lacked these properties – as in many larger scale datasets where high utility speech data is relatively scarce.

\subsection{Summary}
We wanted to test whether we could design a curriculum based on the complexity of the various
pre-training corpora (see Figure~\ref{fig: corpheatmap}). We find that following this approach led to improvements over the reverse, especially on the original set of BLiMP tasks, but failed to show a significant difference over random sampling. We hypothesise that this due to AO-Childes and Open Subtitles, two of the most high utility corpora for grammar acquisition, already making up a large percentage of inputs in the random-sampling setting. Thus, the introduction of a curriculum may have little impact.

\section{Control Dataset}
\label{sec:control}
\begin{table}[]
%\centering
\caption{Control Dataset Statistics}
\resizebox{\columnwidth}{!}{%
\begin{tabular}{@{}lccc@{}}
\toprule
Name       & Tokens \% & Input \% & Curriculum Input \% \\ \midrule
AO-Childes & 4\%    & 15.8\%           & 26\%                        \\
CBT        & 50\%   & 51.8\%           & 56\%                        \\
Wikipedia  & 46\%   & 32.4\%           & 18\%                        \\ \bottomrule
\end{tabular}%
}
\label{tab: control corp}
\end{table}

To test whether complexity ordering helped more when the training data was less optimal, we created a new dataset. It consists of the AO-Childes portion of BabyLM 10M, and the CBT and Wikipedia portions of BabyLM 100M, representing the simplest, middle, and most complex corpora respectively. We set max sequence length to 512 to allow training on as much of the data as possible. Combined, these three corpora have approximately 10 million tokens (similar to the `strict-small' track), but with the vast majority of these now coming from text data. It also means that the number of inputs that come from simpler, more beneficial data is reduced. Descriptive statistics can be found in Table~\ref{tab: control corp}.

\begin{table}[!b]
\centering
\caption{Control Dataset Results on Zero-shot Tasks. Results averaged across 3 seeds.}
\label{tab: controlcres}
\resizebox{\columnwidth}{!}{%
\begin{tabular}{llll}
Training Data & Original & Held Out & Overall\\
\midrule
Simple $\rightarrow$ Complex & \textbf{72.18 $\pm$ 0.88} & \textbf{55.52 $\pm$ 1.08} & \textbf{67.28 $\pm$ 0.52} \\
Random Sampling & 70.77 $\pm$ 0.37 & \textbf{55.88 $\pm$ 1.11} & 66.38 $\pm$ 0.1 \\
\end{tabular}%
}
\end{table}

\begin{figure}[t]
\vspace{-1em}
\centering
\includegraphics[width=\columnwidth]{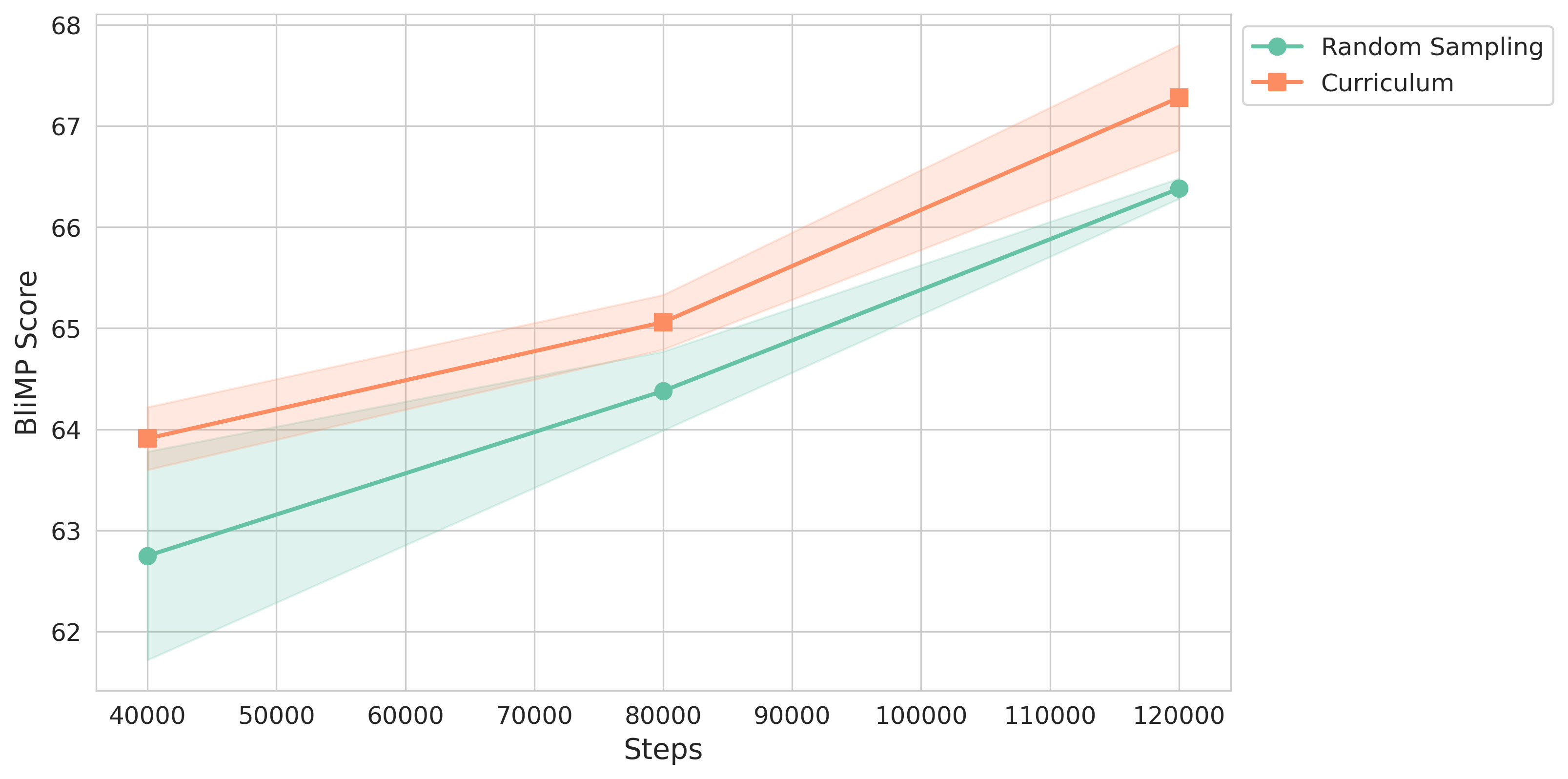}
\caption{Zero-shot performance by step when the model is trained using the curriculum or random sampling on our control dataset (over 3 seeds).}
\label{fig: ccbystep}
\end{figure}

\begin{figure}[]
\centering
\includegraphics[width=\columnwidth]{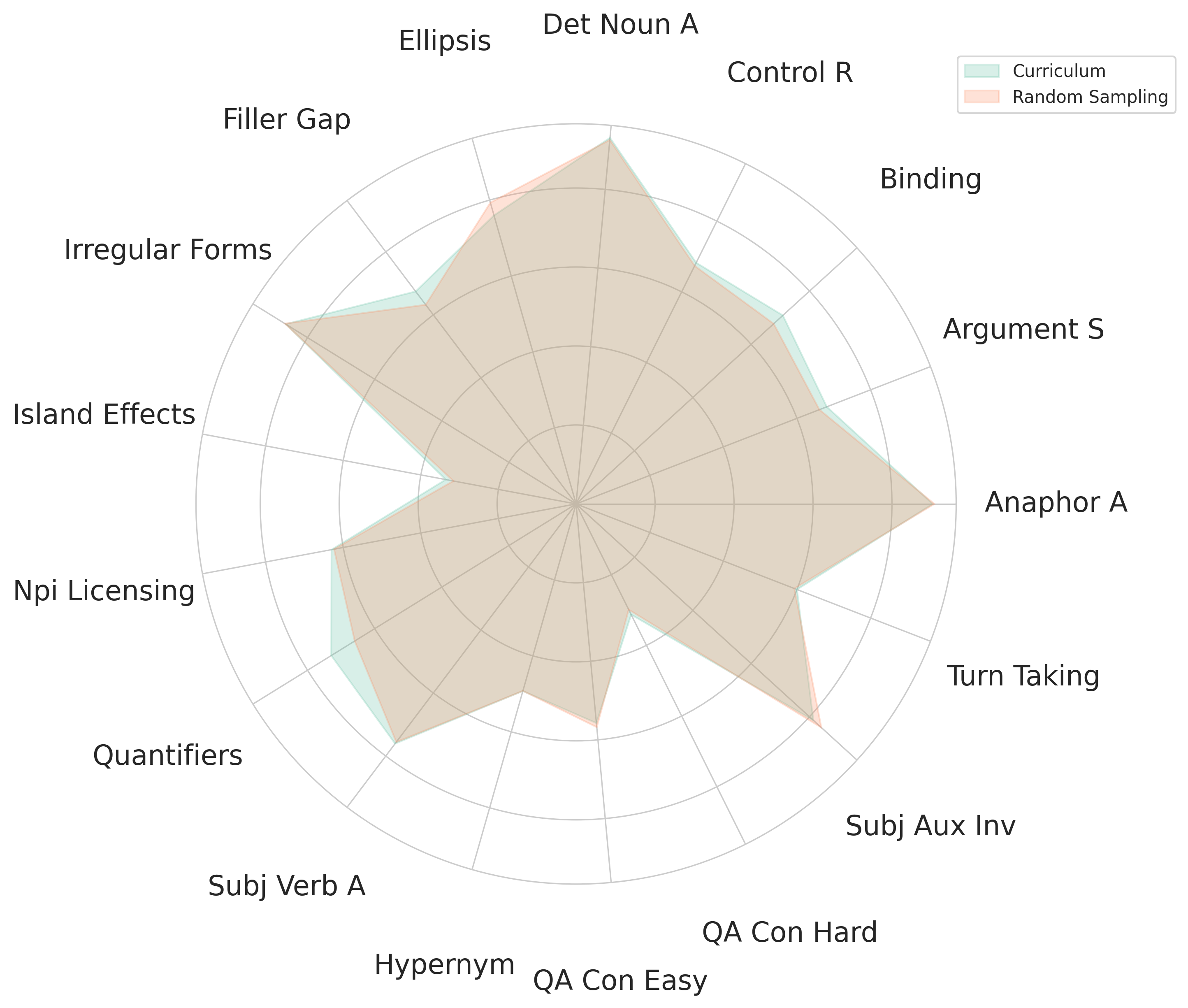}
\caption{By-task breakdown of zero-shot performance on the control dataset curriculum vs random sampling. Results averaged across 3 seeds.}
\label{fig: control}
\end{figure}

We train a new tokeniser on the data, and then compare results between a random sampling baseline and corpus complexity curriculum approach described in the previous section. Both versions are trained for 120k steps, but we had to lower the batch size to 64 due to GPU memory constraints with longer sequences. Results are in Table~\ref{tab: controlcres}, and we plot the by task scores in Figure~\ref{fig: control}. Under this setting, the curriculum approach begins to demonstrate modest but visible improvements over random sampling, though this does not extend to the held out tasks. Figure \ref{fig: ccbystep} shows BLiMP performances as the number of steps increases. The curriculum consistently offers slight improvements over random sampling.

\section{Summary}
We wanted to test whether curriculum learning can be beneficial in a scenario where the majority of data is not as high utility as simple transcribed speech. To do so, we created a control corpus where the majority of data comes from long form text. Under this setting, we find a slight, but discernible improvement from using the curriculum.

\section{Conclusion}
We began our exploration by attempting to design a learning curriculum to further grammar acquisition for the BabyLM strict-small track. We found that when the majority of the data is high-utility, as is the case here, curriculum learning shows no substantial benefits. However, such training data is not always available or may be dwarfed by the number of tokens of low utility data available. In these settings---common for pre-training NLP models---our results indicate some promise in starting small after all. However, extensive further experimentation, most likely requiring larger scale corpora, is necessary to properly test and verify this claim.

\section{Limitations}
The work presented in this paper is our initial foray into starting-small-style learning.
%The following extensions could build upon the work presented here and help shine further light on the nuances of this style of learning.
There are a number of extensions and further questions one could ask, building upon the work presented here, that could help shine further light on the nuances of this style of learning.
% There are a number of additional experiments and ablations that we were not able to finish on time, but are currently running and with which we will augment this paper. Briefly, these are:
\begin{compactitem}
% \item Testing the reverse curriculum for our control dataset. % what could this show? say something along the lines of 'although current results show x, we'd like to know if y holds by running the reverse curriculum.'
\item Although the control-dataset experiments in Section~\ref{sec:control} show better performance when starting small compared to random sampling, we don't yet definitively discount that starting large \emph{in the same setting} does not achieve the same results. This could be remedied by constructing a careful `complementary' large-to-small complexity curriculum.
% \item Training for an increased number of steps for both the random sampling baselines and the corpus curricula, on both the original dataset and the control and reporting performance trends over the course of training. This to measure whether the eventual trend will resemble that reported by \citet{Rohde1999LanguageAI} or that of \citet{BengioC}.
\item Because our training regime is short, for both random sampling and corpus curricula, on both the original data and the control, we don't know whether the eventual trends over training will resemble those reported by \citet{Rohde1999LanguageAI} or \citet{BengioC}. We could explore this by training over longer horizons to see whether comparable trends emerge.
%Explicitly, what are their trends? Which one favors starting small, which favors starting big? Rohde 1999 finds training on complex from the start is better than training on simpler sentences initially, seeming to contradict the "starting small" hypothesis.
\item Our submission for the competition used an additional technique: layer stacking \cite{layerstack}, which progressively grew the model as we advanced through the curriculum (following \citet{Elman1993LearningAD}). The idea was that we would start small in two ways: from simple data and/or a simple model. This yielded some slight improvements over only using the corpora curriculum over the entirety of the strict-small training data, which had been our previous best scoring model.
We do not yet have a complete picture of how layer-stacking affects all the various training regimes discussed in this manuscript, and hence only describe the basic algorithm in the appendix \ref{sec: ls}. % SID: put the layer stacking bit in the appendix with what results you have (avg complexity ranking).
% However, we have not yet completed evaluating the effect of layer stacking on all the various training regimes presented in this paper. Therefore, we chose to omit inclusion of layer stacking from the paper until we are able to make more evidenced based claims on its effect.
\item Follow on work could probe ratios of high utlity versus low utility data to test how much of a token disparity can be tolerated before losing the benefits of starting small from transcribed speech. In our setting, one could replace CBT with a larger proportion of Wikipedia; e.g. Wiki-103 \cite{Wiki-103}.
% It would be interesting to explore how much of a token disparity can be introduced before starting small from transcribed speech loses its benefit.  This could be done, for example, by replacing CBT with a larger proportion of Wikipedia like, e.g., Wiki-103 (Merity et al., 2016).
% SID: explain what the finding is and why training on wiki-103 is a sensible thing here?
%Different data sources will naturally have different distributions of tokens. For example, transcribed speech might contain more informal language, colloquialisms, and speech-specific features (like stutters or filler words), while Wikipedia articles are more formal and polished. A language model trained on a mix of these sources will need to be able to handle this disparity. If the disparity becomes too large, the model might struggle to learn effectively from the mixed data. However, it's difficult to quantify exactly how much disparity is "too much" without empirical testing.
\end{compactitem}

% \clearpage
\begin{table}[!b]
\centering
\caption{Full results from Dynabench for our submission vs. the official RoBERTa baseline for the challenge.}
\label{tab: full}
\scalebox{0.7}{%
\begin{tabular}{@{}lcc@{}}
\toprule
Task & Ours & RoBERTa Base \\
\midrule
Anaphor Agreement & \textbf{84} & 82 \\
Argument Struct & \textbf{70} & 67 \\
Binding & \textbf{69} & 67 \\
Control R & \textbf{70} & 68 \\
DN Agreement & \textbf{92} & 91 \\
Ellipsis & \textbf{77} & 76 \\
Filler Gap & \textbf{76} & 64 \\
Irregular Forms & 87 & 87 \\
Island Effects & \textbf{42} & 40 \\
NPI Licensing & \textbf{65 }& 56 \\
Quantifiers & \textbf{78} & 71 \\
SV Agreement & \textbf{77} & 66 \\
Hypernym & 45 & \textbf{49} \\
QA Cong Easy & \textbf{69} & 31 \\
QA Cong Hard & \textbf{33 }& 32 \\
SA Inversion & \textbf{77} & 72 \\
Turn Taking & \textbf{57} & 53 \\
\midrule
CoLA & \textbf{32} & 26 \\
SST-2 & 87 & 87 \\
MRPC & 79 & 79 \\
QQP & \textbf{82} & 74 \\
MNLI & 73 & 73 \\
MNLI-MM & 74 & 74 \\
QNLI & \textbf{78} & 77 \\
RTE & 49 & \textbf{62} \\
BoolQ & 62 & \textbf{66} \\
MultiRC & 60 & \textbf{61} \\
WSC & 61 & 61 \\
\midrule
CR & \textbf{0.73} & 0.43 \\
LC & 1.0 & 1.0 \\
MV & \textbf{1.0} & 0.98 \\
RP & 0.84 & \textbf{0.94} \\
SC & 0.16 & \textbf{0.86} \\
CR\_LC & -0.58 & \textbf{-0.28} \\
CR\_RTP & -0.92  & \textbf{-0.77} \\
MV\_LC & -1.0 & \textbf{-0.99} \\
MV\_RTP & \textbf{-0.26} & -0.79 \\
SC\_LC & -0.43 & \textbf{0.16} \\
SC\_RP & -0.59 & -0.45 \\
\bottomrule
\end{tabular}
}
\end{table}

\section{Full Results}
While our focus here has been grammar acquisition, we present results on all tasks in Table~\ref{tab: full}. We perform favourably compared to the official RoBERTa baseline for the challenge, but one area shows a notable disparity---MSGS tasks~\cite{MSGS} measuring syntactic category. This may be because our model is too shallow (RoBERTa base has 12 layers vs. our 8).

\section*{Acknowledgements}
MO was funded by a PhD studentship through Huawei-Edinburgh Research Lab Project 10410153. We also wish to thank Victor Prokhorov for his suggestions and tireless willingness to answer questions.

% \clearpage

\bibliography{custom}

\begin{thebibliography}{15}
\expandafter\ifx\csname natexlab\endcsname\relax\def\natexlab#1{#1}\fi

\bibitem[{Bengio et~al.(2009)Bengio, Louradour, Collobert, and
  Weston}]{BengioC}
Yoshua Bengio, J\'{e}r\^{o}me Louradour, Ronan Collobert, and Jason Weston.
  2009.
\newblock \href {https://doi.org/10.1145/1553374.1553380} {Curriculum
  learning}.
\newblock In \emph{Proceedings of the 26th Annual International Conference on
  Machine Learning}, ICML '09, page 41–48, New York, NY, USA. Association for
  Computing Machinery.

\bibitem[{Campos(2021)}]{NoLingCurriculum}
Daniel Campos. 2021.
\newblock \href {http://arxiv.org/abs/2108.02170} {Curriculum learning for
  language modeling}.
\newblock \emph{CoRR}, abs/2108.02170.

\bibitem[{Elman(1993)}]{Elman1993LearningAD}
Jeffrey~L. Elman. 1993.
\newblock \href {https://api.semanticscholar.org/CorpusID:2105042} {Learning
  and development in neural networks: the importance of starting small}.
\newblock \emph{Cognition}, 48:71--99.

\bibitem[{Gong et~al.(2019)Gong, He, Li, Qin, Wang, and Liu}]{layerstack}
Linyuan Gong, Di~He, Zhuohan Li, Tao Qin, Liwei Wang, and Tieyan Liu. 2019.
\newblock \href {https://proceedings.mlr.press/v97/gong19a.html} {Efficient
  training of {BERT} by progressively stacking}.
\newblock In \emph{Proceedings of the 36th International Conference on Machine
  Learning}, volume~97 of \emph{Proceedings of Machine Learning Research},
  pages 2337--2346. PMLR.

\bibitem[{Huebner et~al.(2021)Huebner, Sulem, Cynthia, and Roth}]{babyberta}
Philip~A. Huebner, Elior Sulem, Fisher Cynthia, and Dan Roth. 2021.
\newblock \href {https://doi.org/10.18653/v1/2021.conll-1.49} {{B}aby{BERT}a:
  Learning more grammar with small-scale child-directed language}.
\newblock In \emph{Proceedings of the 25th Conference on Computational Natural
  Language Learning}, pages 624--646, Online. Association for Computational
  Linguistics.

\bibitem[{Liu et~al.(2019)Liu, Ott, Goyal, Du, Joshi, Chen, Levy, Lewis,
  Zettlemoyer, and Stoyanov}]{RoBERTa}
Yinhan Liu, Myle Ott, Naman Goyal, Jingfei Du, Mandar Joshi, Danqi Chen, Omer
  Levy, Mike Lewis, Luke Zettlemoyer, and Veselin Stoyanov. 2019.
\newblock \href {http://arxiv.org/abs/1907.11692} {Roberta: {A} robustly
  optimized {BERT} pretraining approach}.
\newblock \emph{CoRR}, abs/1907.11692.

\bibitem[{Merity et~al.(2016)Merity, Xiong, Bradbury, and Socher}]{Wiki-103}
Stephen Merity, Caiming Xiong, James Bradbury, and Richard Socher. 2016.
\newblock \href {http://arxiv.org/abs/1609.07843} {Pointer sentinel mixture
  models}.
\newblock \emph{CoRR}, abs/1609.07843.

\bibitem[{Mueller and Linzen(2023)}]{PlantTrees}
Aaron Mueller and Tal Linzen. 2023.
\newblock How to plant trees in language models: Data and architectural effects
  on the emergence of syntactic inductive biases.
\newblock \emph{ArXiv}, abs/2305.19905.

\bibitem[{Murty et~al.(2023)Murty, Sharma, Andreas, and
  Manning}]{MurtyGrokking}
Shikhar Murty, Pratyusha Sharma, Jacob Andreas, and Christopher~D. Manning.
  2023.
\newblock Grokking of hierarchical structure in vanilla transformers.
\newblock \emph{ArXiv}, abs/2305.18741.

\bibitem[{Nagatsuka et~al.(2021)Nagatsuka, Broni-Bediako, and
  Atsumi}]{BlockCurriculum}
Koichi Nagatsuka, Clifford Broni-Bediako, and Masayasu Atsumi. 2021.
\newblock \href {https://aclanthology.org/2021.ranlp-1.112} {Pre-training a
  {BERT} with curriculum learning by increasing block-size of input text}.
\newblock In \emph{Proceedings of the International Conference on Recent
  Advances in Natural Language Processing (RANLP 2021)}, pages 989--996, Held
  Online. INCOMA Ltd.

\bibitem[{Rohde and Plaut(1999)}]{Rohde1999LanguageAI}
Douglas L.~T. Rohde and David~C. Plaut. 1999.
\newblock \href {https://api.semanticscholar.org/CorpusID:961169} {Language
  acquisition in the absence of explicit negative evidence: how important is
  starting small?}
\newblock \emph{Cognition}, 72:67--109.

\bibitem[{Surkov et~al.(2022)Surkov, Mosin, and
  Yamshchikov}]{surkov-etal-2022-data}
Maxim Surkov, Vladislav Mosin, and Ivan Yamshchikov. 2022.
\newblock \href {https://doi.org/10.18653/v1/2022.insights-1.16} {Do data-based
  curricula work?}
\newblock In \emph{Proceedings of the Third Workshop on Insights from Negative
  Results in NLP}, pages 119--128, Dublin, Ireland. Association for
  Computational Linguistics.

\bibitem[{Warstadt et~al.(2023)Warstadt, Choshen, Cotterell, Linzen, Mueller,
  Wilcox, Adina, and Zhuang}]{BabyLM}
Alex Warstadt, Leshem Choshen, Ryan Cotterell, Tal Linzen, Aaron Mueller, Ethan
  Wilcox, Williams Adina, and Chengxu Zhuang. 2023.
\newblock Findings of the {B}aby{LM} {C}hallenge: {S}ample-efficient
  pretraining on developmentally plausible corpora.
\newblock In \emph{Proceedings of the {B}aby{LM} {C}hallenge}. Association for
  Computational Linguistics (ACL).

\bibitem[{Warstadt et~al.(2020)Warstadt, Zhang, Li, Liu, and Bowman}]{MSGS}
Alex Warstadt, Yian Zhang, Xiaocheng Li, Haokun Liu, and Samuel~R. Bowman.
  2020.
\newblock \href {https://doi.org/10.18653/v1/2020.emnlp-main.16} {Learning
  which features matter: {R}o{BERT}a acquires a preference for linguistic
  generalizations (eventually)}.
\newblock In \emph{Proceedings of the 2020 Conference on Empirical Methods in
  Natural Language Processing (EMNLP)}, pages 217--235, Online. Association for
  Computational Linguistics.

\bibitem[{Wolf et~al.(2019)Wolf, Debut, Sanh, Chaumond, Delangue, Moi, Cistac,
  Rault, Louf, Funtowicz, and Brew}]{transformers}
Thomas Wolf, Lysandre Debut, Victor Sanh, Julien Chaumond, Clement Delangue,
  Anthony Moi, Pierric Cistac, Tim Rault, R{\'{e}}mi Louf, Morgan Funtowicz,
  and Jamie Brew. 2019.
\newblock \href {http://arxiv.org/abs/1910.03771} {Huggingface's transformers:
  State-of-the-art natural language processing}.
\newblock \emph{CoRR}, abs/1910.03771.

\end{thebibliography}
\bibliographystyle{acl_natbib}

\clearpage

\appendix
%The appendices could do with a look. Should probably have the Curriculum Stages figure in App A. Table for B should probably be on the same page as App B heading.
\section*{Appendices}
\section{Layer Stacking}
\label{sec: ls}
We grew our model during training by adding a layer when we reached each new stage of our curriculum. We cloned the existing uppermost layer at the beginning of each new stage of our curriculum, then stacked that layer on top of the existing layers of our model. Our model then proceeds to learn from our new mix of datasets for the new stage of the curriculum, with the uppermost layer most responsive to the newly revealed datasets in our curriculum. In this way we progressed from 1 to 8 layers over the course of our training regime.
\begin{figure}[h]
\centering
\includegraphics[width=\columnwidth]{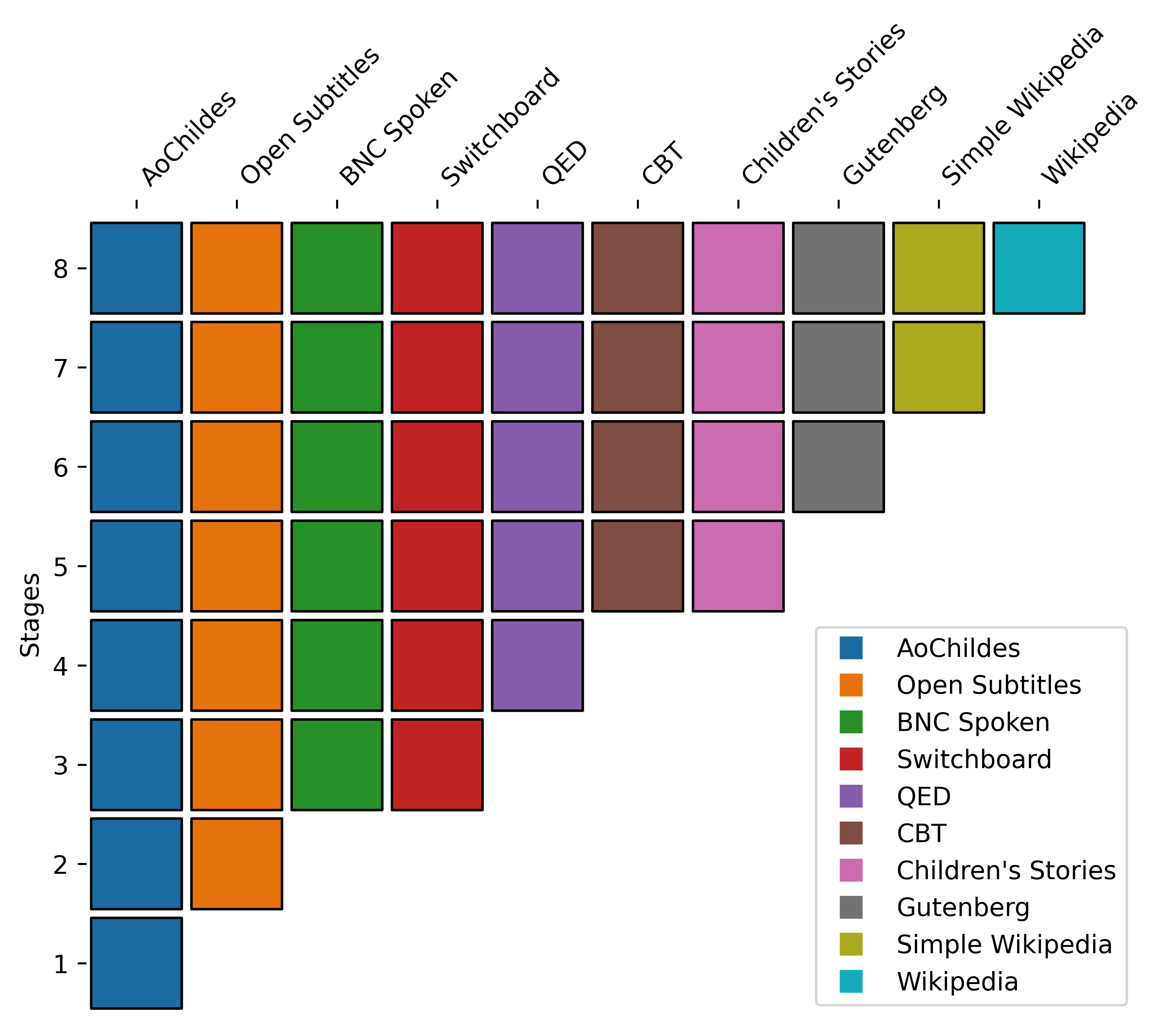}
\caption{Our learning curriculum exposes our model to additional datasets stage-by-stage as it progresses through our training regime.}
\label{fig: curriculum}
\end{figure}
\section{Full Results Table}
%!But we had a Full Results section in the main text already?
% \onecolumn
\begin{table*}[hb]
\centering
\caption{Sequence vs block input performance on zero-shot tasks. Results are averaged across three random seeds. \ref{fig: seqvblock}}
\label{tab:seqvblocfull}
%\resizebox{\textwidth}{!}{%
%\begin{tabular}{@{}lllllllllllllllllll@{}}
%Model & Anaphor A & Argument S & Binding & Control R & Det Noun A & Ellipsis & Filler Gap & Irregular Forms & Island Effects & Npi Licensing & Quantifiers & Subj Verb A & Hypernym & QA Con Easy & QA Con hard & Subj Aux Inv & %Turn Taking & Score \\ \midrule
%Sequences & \textbf{86.78} & \textbf{70.84} & \textbf{68.58} & \textbf{66.60} & \textbf{92.31} & \textbf{74.12} & \textbf{74.20} & \textbf{89.47} & \textbf{42.91} & \textbf{58.97} & \textbf{76.06} & \textbf{76.52} & \textbf{49.19} & \textbf{65.63} & 29.90 & \textbf{81.21} & 56.31 & \textbf{68.21} \\
%Blocks & 86.39 & 63.59 & 65.36 & 63.66 & 80.91 & 73.81 & 67.19 & 77.89 & 39.08 & 43.85 & 68.20 & 61.81 & 48.22 & 64.06 &\textbf{49.50} & 76.99 & \textbf{59.17} & 64.01 \\
%\end{tabular}%
%}
%App.B: Table on subsequent page by mistake.
\resizebox{\textwidth}{!}{%
\begin{tabular}{@{}lccccccccc@{}}
Model & Anaphor A & Argument S & Binding & Control R & Det Noun A & Ellipsis & Filler Gap & Irregular Forms & Island Effects \\ \midrule
Sequences & \textbf{86.78} & \textbf{70.84} & \textbf{68.58} & \textbf{66.60} & \textbf{92.31} & \textbf{74.12} & \textbf{74.20} & \textbf{89.47} & \textbf{42.91} \\
Blocks & 86.39 & 63.59 & 65.36 & 63.66 & 80.91 & 73.81 & 67.19 & 77.89 & 39.08 \\[2ex]
Model & Npi Licensing & Quantifiers & Subj Verb A & Hypernym & QA Con Easy & QA Con hard & Subj Aux Inv & Turn Taking & \textbf{Score} \\ \midrule
Sequences & \textbf{58.97} & \textbf{76.06} & \textbf{76.52} & \textbf{49.19} & \textbf{65.63} & 29.90 & \textbf{81.21} & 56.31 & \textbf{68.21} \\
Blocks & 43.85 & 68.20 & 61.81 & 48.22 & 64.06 &\textbf{49.50} & 76.99 & \textbf{59.17} & 64.01 \\
\end{tabular}%
}

\end{table*}

\end{document}